# Robust Mobile Robot Path Planning via LLM-Based Dynamic Waypoint Generation

Muhammad Taha Tariq[a], Yasir Hussain[b,*] and Congqing Wang[a,*]

[a]*College of Automation Engineering, Nanjing University of Aeronautics and Astronautics, , Nanjing, 210016, China*
[b]*School of Computer Science and Mathematics, Liverpool John Moores University, Byrom Street, Liverpool, L3 3AF, UK*

ABSTRACT

Mobile robot path planning in complex environments remains a significant challenge, especially in achieving efficient, safe and robust paths. The traditional path planning techniques like DRL models typically trained for a given configuration of the starting point and target positions, these models only perform well when these conditions are satisfied. In this paper, we proposed a novel path planning framework that embeds Large Language Models to empower mobile robots with the capability of dynamically interpreting natural language commands and autonomously generating efficient, collision-free navigation paths. The proposed framework uses LLMs to translate high-level user inputs into actionable waypoints while dynamically adjusting paths in response to obstacles. We experimentally evaluated our proposed LLM-based approach across three different environments of progressive complexity, showing the robustness of our approach with llama3.1 model that outperformed other LLM models in path planning time, waypoint generation success rate, and collision avoidance. This underlines the promising contribution of LLMs for enhancing the capability of mobile robots, especially when their operation involves complex decisions in large and complex environments. Our framework has provided safer, more reliable navigation systems and opened a new direction for the future research. The source code of this work is publicly available on GitHub[1].

## 1. INTRODUCTION

Large Language Models, including Open Ai, Gemini and Ollama, are revolutionizing the way for mobile robots to communicate with humans and plan their trajectory in complex environments Sun et al. (2024), Shah et al. (2023). Although the environments which are static with some obstacles like walls, stairs, windows and tables etc. The power of LLMs to understand natural language commands Patki et al. (2020), Roy et al. (2019) and the easy integration into real-time decision-making Arkin et al. (2020), Barber et al. (2016) processes makes them quite suitable for handling mobile robot path planning and navigational tasks. In this section, we discussed the potential of LLMs in such environments, with an emphasis on real-time replanning around obstacles. Modern large language model such as GPT-4 Achiam et al. (2023), Gemini 1.5 Flash Team et al. (2023) and llama 3.2 Touvron et al. (2023) are designed to understand the natural languages and to generate a text-based output on the base of prompts that provided by users Guan et al. (2023), Huang et al. (2023), Kim et al. (2024). These models are mainly useful for the tasks where question answering, cybersecurity Alturkistani and Chuprat (2024), and language translations are required. But in the context of mobile robots, these LLMs are now being used to interpret the human instructions and convert them into desirable tasks like generating robot programs Pu et al. (2024), Virtual Human Llanes-Jurado et al. (2024), fault detection Baghernezhad (2012), and image understandings Wang et al. (2024).

Among the most compelling reasons LLMs are sought in mobile robot path planning, especially under varying environmental conditions with changing starting and target positions, is because they are more robust and flexible compared to DRL-based approaches. Whereas the DRL models are trained for any given particular task Zhu and Zhang (2021) and hence show limited adaptability when changing the starting or target position of the robot, LLMs are pre-trained Brown (2020) over a large-scale dataset and thus will readily take in dynamic inputs. The DRL models are typically trained for a given configuration of the starting and target positions; these models only perform well when these conditions are satisfied Biagiola and Tonella (2024). In case the environment changes, such as the starting location or the target position changed, the model may need to be retrained or fine-tuned to maintain the best performance in that environment Zhao et al. (2024). On the other hand, LLMs can generate strong, context-dependent paths for navigation without retraining. Given their ability to process natural language commands and their ease of adaptation to new spatial contexts, LLMs provide a very flexible solution for path planning tasks in real time when either the starting point or the target object may change often. This advantage makes the LLM-based path planning especially well-suited for environments requiring flexibility, adaptability, and human-robot interaction without the overhead of extensive retraining as demanded by DRL systems.

---

[1]`https://github.com/DC1one/LLM-Based-mobile-robot-path-planning`

*Corresponding authors

✉ mtuaha@nuaa.edu.cn (M.T. Tariq); yaxirhuxxain@yahoo.com (Y. Hussain); cqwang@nuaa.edu.cn (C. Wang)
ORCID(s):

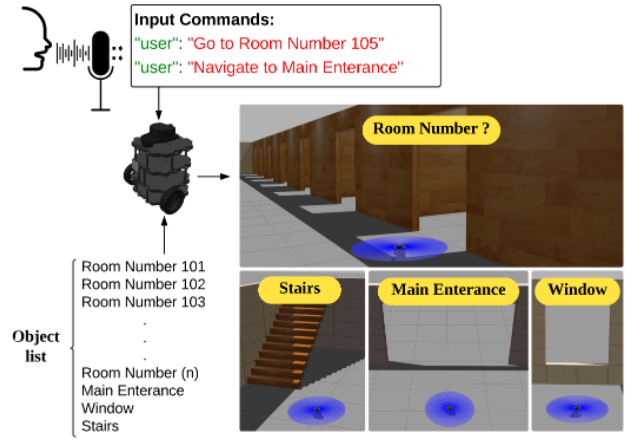

**Figure 1:** Overview of the role of Large Language Models (LLMs) in mobile robot path planning. This figure highlights how LLMs can interpret natural language commands and path planning for mobile robots in complex environments.

LLMs are capable of understanding high-level commands in natural language, such as "Go to room X or Go to Window," and translating such commands into actions for the mobile robot to navigate towards the desire object. The LLMs understand these commands and integrate information with respect to the environment, then they can generate a sequence of waypoints Latif (2024) for mobile robots to reach the desired object while navigating through the environment. According to the current status of research on fully LLM-based mobile robot path planning techniques, most environments are static Zeng et al. (2022), Kannan et al. (2023) but real-time replanning Song et al. (2023) becomes essential for advancing the field to tackle the challenges of avoidance from obstacles when any unexpected obstacles appear in the planned path—for example, a previously or a newly generated sequence of waypoints have a wall or an object in the way of the planned path. In that case to avoid the obstacle and to reach the desire location, the mobile robot requires a real-time replanning mechanism that can generate a new waypoint sequence to helping in reroute the path around the obstruction. The main focus of this research is to make a valuable addition in the field of mobile robot path planning by integrating Large Language Models (LLMs) to make a novel approach (see Fig. 1). This approach addressed with different challenges to improve performance, flexibility, and robust solution for complex environment layouts. Following are the contribution of our proposed approach:

- We developed a novel path planning approach that utilizes a Large Language Model to interpret natural language commands to generate a sequence of waypoint that helps the mobile robot to reach the target object.

- We implemented a robust real time obstacle avoidance and replanning mechanism where robot dynamically detects the obstacles while navigation and replan its path towards the target object with safety and efficiency.

- We integrated a voice command system that process natural language commands from users and translate them into navigational waypoints to enhance the human robot interaction.

With these contributions we are providing robust solutions and equipping the mobile robot with capabilities of real time adaptability, human interactions, safety, efficiency, and versatility in complex environments. The insights gained from this research opened a new avenue to explore advanced AI-driven path planning mobile robot frameworks.

## 2. RELATED WORK

In the broader field of human-robot interaction over the last years, major emphasis has been placed on equipping robots with the ability of understanding and act upon natural language commands Bonarini (2020). Context identification and interpretation by the robot during interactions have emerged as an essential element for any successful interaction between human and robot. Several studies focused on how context helps a robot to perform a task. Some works, such as Kritsotakis et al. (2008) and Carton et al. (2017), studied the use of contextual information by robots to safely navigate pedestrians in indoor settings. In contrast, other work has focused on interactive contexts, where there is a direct human-robot communication are used to translate natural language commands into robotic actions without cumbersome manual programming Carlson and Demiris (2008). Furthermore, the works of Calisi et al. (2007) and Mathews et al. (2009) have used the Simultaneous Localization and Mapping techniques for better navigation of the robot by inferring the environmental features dynamically while avoiding the need for explicit goal specification through enhanced context recognition.

When it is regarding the integration of LLMs for the execution of robotic tasks, recent times have seen quite improved results regarding how robots interpret natural language commands and execute a set of actions based on it. Several works present that complex instructions could also be processed by LLMs and converted into executable ones. For Instance, SayCan Ahn et al. (2022) demonstrated that LLMs can use semantic knowledge and pre-training to perform tasks described by a natural language specification; while the application of Open AI's ChatGPT on robotic control Vemprala et al. (2024), from drone navigation to robotic manipulation, uses effective prompt engineering coupled with API integrations. Other notable frameworks, such as NavCon Biggie et al. (2023), serve as an intermediary interface between LLMs and robotic navigation systems. In this approach it combines the natural language input along with visual data and map-based information to generate Python code that executes navigation tasks. The idea is further extended in VLMaps Huang et al. (2023) to integrate visual-language models with 3D map representations, which enable the robot to create waypoints and navigate grounded on natural language. However, the limitation of VLMaps lies in the fact that it relies on visual inputs, which are pretty challenging in scenarios where visual data is too poor or ambiguous. Another approach, PROGPROMPT Singh et al. (2023), indicated how LLMs could be applied to task planning by embedding structured prompts into the system to enhance its generalization ability. In parallel, other frameworks like LLM-DP Dagan et al. (2023) and CAPEAM Kim et al. (2023) have worked on context-aware planning that allows LLMs to adapt to different input scenarios and change the navigation strategy of the robot. These systems improve task execution by incorporating context-sensitive adjustments based on the environment and the nature of the particular task.

Despite these advances, one of the big challenges involves translating complex, often ambiguous, and large sets of natural language commands into precise executable navigation waypoints. Our proposed framework presents LLM-based method with a much robust obstacle avoidance system tailored for complex environments. In this approach, we focused on the real-time generation of pathways and navigation optimization, making the interpretation of user commands highly accurate, while it can adaptively plan the routes to reach specified targets. Further, we enhance this with integrated voice command functionality to make human-robot interactions seamless, relevant to the context, and actionable. In the following this article is organized where, Section 3 presents and analyses the proposed LLM-based path planning framework, outlining its architecture and methodology. In Section 4, we describe the experimental setup and the design of test environments. Section 5 presents a comprehensive discussion on the results mainly focusing on key performances, we also addressed the mitigation strategies for system limitations and future work at the end. The last Section 6 concludes the article and summarizing the findings.

## 3. PROPOSED FRAMEWORK

In this section, we elaborate on how the proposed method is designed for integrating Large Language Model (LLM) capabilities into a mobile robot path planning system. We will describe the methodology of waypoint generation that how it generates, validate, and execute these waypoints for navigation of a mobile robots in pre-defined environments. This process is subdivided into several stages: a Waypoint Generation Process, which is a process whereby the system translates high-level user instructions into a set of waypoints; and secondly, Waypoint Parsing and Validation, where the generated waypoints must be checked against viability under the environmental constraints at corridor boundaries and safe margins. Next to this, we discuss Navigation Strategies, optimizing the generation of a path; then, details on how this course is executed and controlled for guiding the robot's movement are provided. We also cover Obstacle Detection and Emergency Handling, explaining how the system dynamically adapts to changes in the environment in real-time. We finally cover the Replanning Requests at the end, how the proposed system can efficiently handle path adjustments, followed by a discussion of Navigation States and Transitions that govern the robot's operational phases.

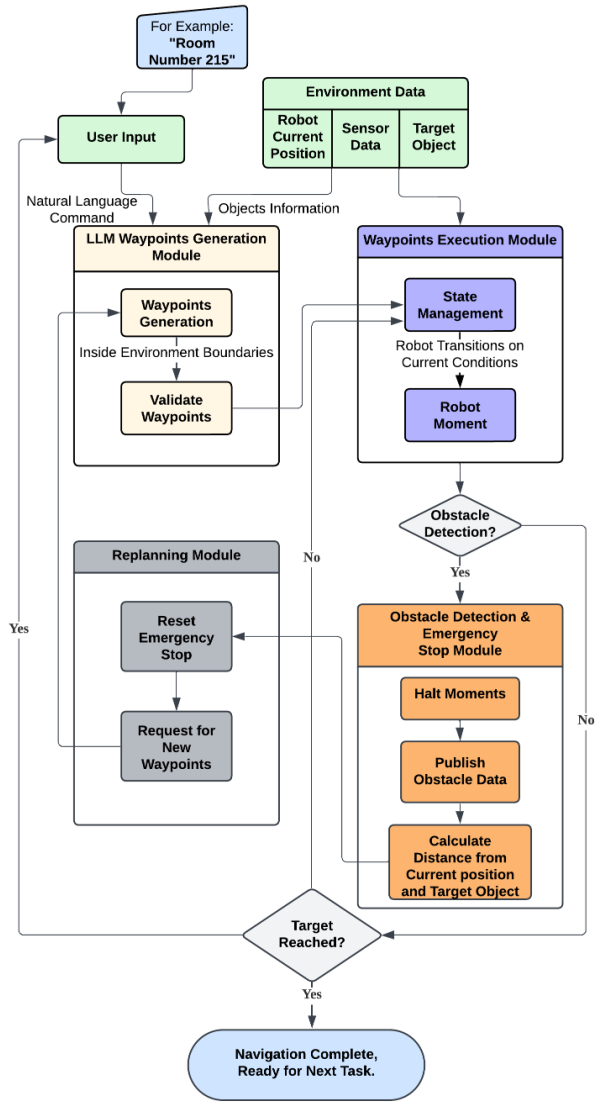

**Figure 2:** Architecture of proposed framework that integrates large language models capabilities into path planning for mobile robots. The figure shows the flow from the user commands through to the generation and execution of waypoints, with particular emphasis on the role of LLM in interpreting these instructions and enabling real-time navigation.

Our framework takes advantage of the strengths of the Ollama LLMs in the translation of natural language commands into executable navigation tasks. This way, by letting LLMs be incorporated into the path-planning process, the system could understand high-level user instructions and adapt its navigation strategies in real-time. The proposed system will provide seamless communication between the human operators and the robots, hence enhancing the general efficiency and flexibility of the mobile robot path planning. This framework is designed to handle environmental conditions, such as predefined obstacles, and real-time replanning requirements by unexpected obstructions between the planned path in the environment. The architecture of our proposed framework showing major components and information flows between the LLM, user inputs, and the robot navigation system in Fig. 2. This integration of the Ollama Natural language Processing (NLP) capabilities with a path-planning system will help in complex large environments, multi-step commands and support of changes in the robot trajectory.

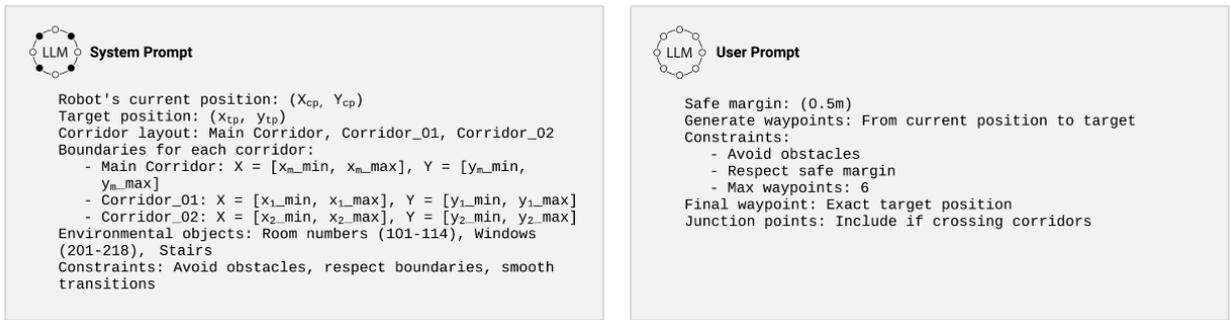

**Figure 3:** System and User Prompt Interaction for Waypoint Generation: An example of prompt engineering, communication between the system prompt and the user prompt, where the system prompt defines the environmental parameters, robot position, and the constraints; on the other hand, the user prompt specifies the task and requirements for path planning. This interaction forms the basis for effective waypoints generation in robotic navigation.

The key role in enabling natural language processing for path planning of the mobile robot is integration by the Ollama LLM models. Ollama, which is an open-source large language model library, is characterized by high flexibility and easy integrability. It does not require external API keys and thus loads the language model directly onto the control system of a robot for perfect interaction without any third-party service dependence. When the user gives a voice command to robot, for example, "Go to Corridor 2", the robot first captures the spoken input through a speech recognition system like Google's Speech Recognition Deuerlein et al. (2021) API that we used in this framework. Then the audio is converted into text, which is sent to the Ollama model for interpretation. Ollama processes the text and generates relevant navigation instructions that the robot will use to generate waypoints and execute path planning. The data flow in this integration is straightforward: the user's voice command is processed by the speech recognition system and then passed to Ollama for interpretation. Then the LLM outputs an actionable navigation JSON array of coordinates, which are sent to the robot's control system for execution of the planned path. This system thus empowers the robot to follow user instructions in real time and handle different navigation challenges efficiently.

### 3.1. Prompt Engineering and Natural Language Processing

This system includes Prompt Engineering and Natural Language Processing in a section where the commands given by the user has to be understood and translated into path-planning instructions. So when the user gives a command, for example, "Go to Room Number 101" or "Go to the window," first of all, speech recognition will convert the audio input into text. The textual command is then parsed to extract relevant information, such as the target room or object. Prompt Engineering ensures that the recognized command is in a format that the system understands. It does this by finding the target object, such as a room, window, or any other object, using regular expressions. The command "go to Room Number 101" is parsed to find "Room-number-plate-101" as the target object from a predefined list of objects in the environment. Upon such information extraction of the target, the system passes it on to the NLP model for further interpretation of the command and generation of the path toward the target object. Path planning considers the current position of the robot and the coordinates of the object, retrieved from the object list of the environment. The result is a set of waypoints guiding the robot from its current location to the target. This proposed system parses simple commands and outputs appropriate movement instructions for the robot by this language interpretation process. In our approach, we designed System and User Prompts that work together to guide the LLM in generating accurate and effective navigation paths. The System Prompt provides the LLM with critical environmental information, including corridor dimensions, the current position of the robot, and safe margins from walls. Such data defines the spatial context; hence, the model knows the surroundings and constraints of the robot before making a decision. The User Prompt, as in the following example, is the translation of the task to a specific request, which tells the LLM how to generate a sequence of waypoints for the robot. It specifies that the robot should move from its current position to a target while keeping safe margins and avoiding obstacles. The User Prompt also sets additional parameters: it limits the number of waypoints according to the environment as we set to 6 waypoints in environment (c), other environments have different limits for waypoints depending on the agent is operating in how much larger environment.

The integration of both prompts is useful for two reasons. First, the System Prompt provides the model with a contextual understanding of all the environmental factors involved in path generation; hence, putting all the constraints at the beginning improves the model's decision-making and avoids ambiguous or inefficient paths. This structured input allows the LLM to generate routes optimized for robust navigation through complex corridors as illustrated in Fig. 3. When these two prompts are used in conjugation, it enhances system performance since there is less likelihood of error and consistency in navigation. The User Prompt narrows down what action the LLM should take based on the context provided by the System Prompt, resulting in more accurate, efficient, and safe navigation plans.

### 3.2. Waypoints Generation

The waypoints are generated using a systematic approach for the mobile robots to navigate with safety and efficiency in the environments. The environment is defined by a single or by three distinct corridors, as shown in Fig. 5, as in environment (c) map, each corridor having its own geometric constraints, as well as a safe margin to ensure the robot maintains a sufficient distance from obstacles. The main goal is to generate a sequence of waypoints that drives the robot from its current position to the target position, respecting the boundaries of these corridors and avoiding obstacles. Let the environment consist of three corridors: Main Corridor, Corridor 01, and Corridor 02. These corridors are designed in the shape of rectangular regions, the length and width of each corridor has a unique set of spatial boundaries defined as follows:

$$x_{\min} \leq x \leq x_{\max}, \quad y_{\min} \leq y \leq y_{\max} \tag{1}$$

here, $x_{\min}$ and $x_{\max}$ are the minimum and maximum x-coordinates limits of the corridor, the robot must have to stay within these defined boundaries limits while navigating in the environment with a safe margin $m$ of 0.5m for maintain a buffer zone to avoid collisions with the corridor walls and this rule is same for y-coordinates. For each corridor $k \in \{\text{Main Corridor}, \text{Corridor 01}, \text{Corridor 02}\}$, the spatial boundaries are defined as follow:

$$\text{Corridor}_k : \left[x_k^{\min}, x_k^{\max}\right], \left[y_k^{\min}, y_k^{\max}\right] \tag{2}$$

Given the current position $(x_{cp}, y_{cp})$ of the robot and the target position $(x_{tp}, y_{tp})$, the aim of the generation process is to generate a set of waypoints $(x_1, y_1), (x_2, y_2), \ldots, (x_n, y_n)$, where each waypoint represents a coordinate that the robot should follow in order to reach a target. The main objectives of the generation process are:

- All waypoints must lie within the corridor boundaries.

- The waypoints shall be spaced with a defined value of 0.7m, such that no two waypoints are redundant or too close to each other.

- The last waypoint must coincide with the target position within a tolerance with a defined value of 0.05m, such that:

$$\sqrt{(x_n - x_{tp})^2 + (y_n - y_{tp})^2} \leq t \tag{3}$$

The waypoints are created in an iterative manner in which the robot moves in discrete steps toward the target. For every step, it computes the next waypoint, taking into account the corridor boundaries and the safe margin. A navigation strategy is employed to decide between a direct line path or a curved path. However, in some cases if the target object and robot are in different corridors then robot needs to determine the current and target corridors based on the robot's current position and the target object position then a junction point must be determined where the robot will transition from one corridor to another, elaborated in Fig. 4.

### 3.3. Waypoint Parsing and Validation

The initial set of waypoints generated has to be validated in order to assure that they fall within the boundaries of the corridors and respect the safety margins. Let the set of generated waypoints be $W = \{(x_1, y_1), (x_2, y_2), \ldots, (x_n, y_n)\}$, the example of generated waypoints which is a JSON array of coordinates as follow:

$$\begin{bmatrix} \{\texttt{"x"}: 1.25, \texttt{"y"}: 2.63\}, \\ \{\texttt{"x"}: 1.25, \texttt{"y"}: 5.36\}, \\ \{\texttt{"x"}: 1.25, \texttt{"y"}: 7.86\} \end{bmatrix} \tag{4}$$

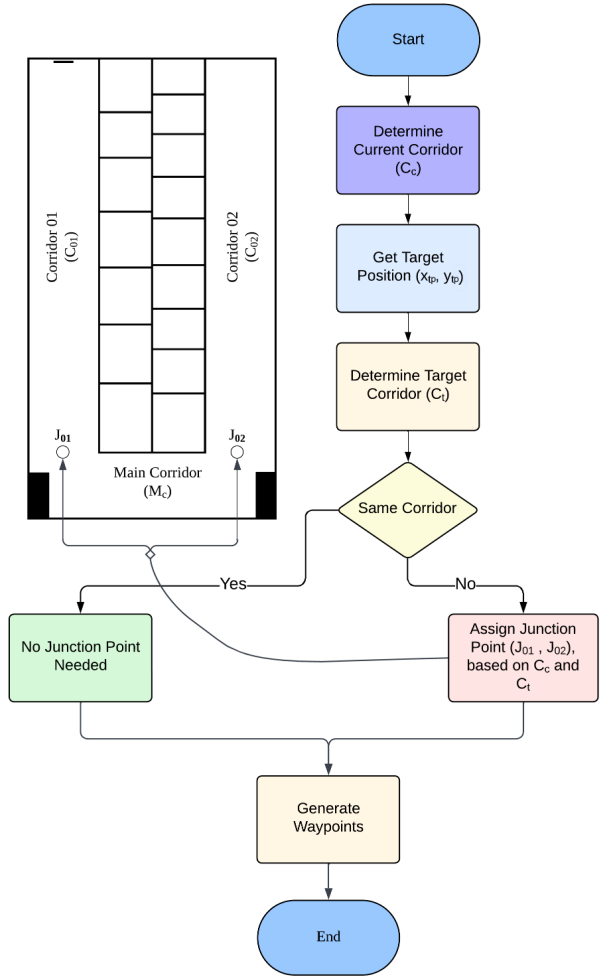

**Figure 4**: The process of generating waypoints for mobile robot navigation. The figure shows the spatial boundaries of the corridor and an iterative approach to generate waypoints from the current position to the target position. During the navigation between corridors, it computes junction points to make transitions smoother and ensures that all the generated waypoints lie within the boundary of the corridors and also are optimally spaced to perform efficient path planning.

For every waypoint $(x_i, y_i) \in W$, we check it first whether it is inside any of the corridors or not. For a point to be considered valid waypoints, the following condition must hold:

$$x_k^{\min} + m \leq x_i \leq x_k^{\max} - m$$
$$\text{and} \quad y_k^{\min} + m \leq y_i \leq y_k^{\max} - m \tag{5}$$

If any waypoint does not satisfy these conditions that waypoint will be discarded from the list, and the last waypoint in the sequence should be the same or as close as possible to the target position $(x_{tp}, y_{tp})$. To ensure that the final waypoint is within an acceptable tolerance, we compute the Euclidean distance $d$ between the final waypoint $(x_n, y_n)$ and the target position:

$$d = \sqrt{(x_n - x_{tp})^2 + (y_n - y_{tp})^2} \tag{6}$$

If this distance value $d$ is greater than the predefined tolerance value $t$, then the waypoints are regarded as invalid and the whole process starts again to generate new waypoints. This is done by setting a small value for the tolerance $t$ (typically, 0.05 m) to ensure accurate arrival at the target object. After successfully generating and validation of

Tariq et al.: *Preprint submitted to Elsevier* Page 7 of 18

waypoints, the final sequence of waypoints is returned such that the robot is able to move from its current position to the target object within the designated corridors and respecting the safe margin. The following are the waypoints as a list of coordinates:

$$W_{\text{final}} = \{(x_1, y_1), (x_2, y_2), \ldots, (x_n, y_n)\} \tag{7}$$

These waypoints are then utilized by the robot's navigation system in following the path to the target object. Before start execution of navigation on these waypoints they plotted out with the corridors of the environment to visually verify the robot's path. Fig. 6 presents the plots of the waypoints and the environment form our experiments; it shows how the robot will navigate from its start position to the target object.

### 3.4. Execution and Control of Waypoints

Waypoints are the predefined target locations or positions that guide the mobile robot from a starting position to reach a desire point in the environment. These are critical element of path planning and usually provided by high-level planners, in our case it is an LLM-based planner. Execution and controlling of these waypoints include receiving, transforming and manage during the navigation of robot in the environment. Once the LLM-based planner provides these waypoints, they are usually publishing in the map frame, which is a global reference frame of the environment. However, the robot operates in the local frame which known as odometry frame. The first step of execution of these waypoints is to transform every waypoint form map frame to the odometry frame of the robot. The waypoints in the map frame represent as $P_{\text{m}} = (x_{\text{m}}, y_{\text{m}})$, which needs to transform into the robot's local odometry frame by applying rotation and translation between these two frames. The robot's position and orientation in the map frame are represented by $(x_{\text{ro}}, y_{\text{ro}}, \theta_{\text{ro}})$, where $\theta_{\text{ro}}$ is the robot's orientation in radians. Then the transformation from the map frame to the odometry frame is given by the following equations:

$$\begin{bmatrix} x_{\text{od}} \\ y_{\text{od}} \end{bmatrix} = \begin{bmatrix} \cos(\theta_{\text{ro}}) & -\sin(\theta_{\text{ro}}) \\ \sin(\theta_{\text{ro}}) & \cos(\theta_{\text{ro}}) \end{bmatrix} \begin{bmatrix} x_{\text{m}} \\ y_{\text{m}} \end{bmatrix} + \begin{bmatrix} x_{\text{ro}} \\ y_{\text{ro}} \end{bmatrix} \tag{8}$$

This equation rotates the coordinates of the waypoint by the orientation of the robot and then translates it by the position of the robot in the map. The waypoint $P_{\text{od}} = (x_{\text{od}}, y_{\text{od}})$ is ready to be executed after transformation. Once the waypoints are prepared, they will be stored in the system for execution. The robot tracks the index of the current waypoint, which reflects the waypoint the robot shall drive towards. The index of the waypoints begins with zero, referring to the first waypoint. When the robot reaches a particular waypoint, the index increases, meaning it would then go ahead towards the next waypoint. It keeps navigating through the list of waypoints in order, correcting position and orientation to follow the path. This continues until all waypoints are completed or until such time as the navigation may be interrupted due to unforeseen circumstances, such as an unexpected obstacle between these waypoints.

### 3.5. Movement Control and Navigation Execution

The movement control and navigation execution are responsible for guiding the robot from one waypoint to the next. This includes calculating the desired velocities of the robot which are linear and angular, and sending appropriate commands for its movement. The motion is to be constrained based on the current pose of the robot, the target waypoint, and any other external factors such as obstacles. The whole navigation is done in steps, whereby at each step the robot decides how it should move towards the current waypoint. The robot calculates two main quantities, which are the distance to the target waypoint and the angle between its current heading and the direction to the waypoint. The Euclidean distance $d$ from the robot's current position $(x_{\text{cp}}, y_{\text{cp}})$ to the target waypoint $(x_{\text{tw}}, y_{\text{tw}})$ is given by:

$$d = \sqrt{(x_{\text{tw}} - x_{\text{cp}})^2 + (y_{\text{tw}} - y_{\text{cp}})^2} \tag{9}$$

The robot also needs to compute the angular error $\alpha$, which is the difference between the robot's current orientation $\theta_{\text{ro}}$ and the angle to the target waypoint. The angle to the waypoint is computed using the atan2 function, which is following:

$$\alpha = \text{atan2}(y_{\text{tw}} - y_{\text{ro}}, x_{\text{tw}} - x_{\text{ro}}) - \theta_{\text{ro}} \tag{10}$$

The angle $\alpha$ is then normalized to the range $[-\pi, \pi]$ to avoid large jumps in orientation. This ensures that the robot can rotate smoothly to face the target without sudden direction changes. Once the distance and angle are calculated,

Tariq et al.: *Preprint submitted to Elsevier* Page 8 of 18

the robot adjusts its velocity according to a proportional control law:

$$V_{\text{linear}} = k_{\text{linear}} \cdot d \tag{11}$$

$$V_{\text{angular}} = k_{\text{angular}} \cdot \alpha \tag{12}$$

where $k_{\text{linear}}$ and $k_{\text{angular}}$ are proportional gain constants. These constants determine how quickly the robot reacts to the error in position and orientation. The linear velocity $V_{\text{linear}}$ is proportional to the distance to the waypoint, meaning that the robot moves faster when it is far from the target and slows down as it approaches. On the other hand, angular velocity $V_{\text{angular}}$ is proportional to the angular error, meaning the robot rotates faster when the orientation error is large. Through these velocities robot adjusts in real-time to move towards the target waypoint, correcting its position and orientation as needed.

### 3.6. Obstacle Detection and Emergency Handling

While navigating in the environment the robot has to avoid the obstacles in its path as well. Our robot is equipped with a LIDAR sensor that provides distance measurements of obstacles. For each laser scan range $r_i$, if the distance falls below the critical distance $d_{\text{critical}}$ of 0.5m, then the robot needs to stop to avoid collisions. If an obstacle is detected in front of the robot, it goes into the emergency stop state. when the measurement of the distance to the closest obstacle is less than the critical threshold of 0.35m then this state becomes activated. After that the robot stops by setting both its linear and angular velocities to zero:

$$V_{\text{linear}} = 0, \quad V_{\text{angular}} = 0 \tag{13}$$

The robot also senses the environment to see whether there is an obstacle exactly in its front within a certain angular range of 15 degrees, usually inside $\pm\theta_{\text{range}}$ degrees. If an obstacle is detected within this range and critical distance, it is considered by the system as an urgent situation that needs replanning. The mathematical formulation for this goes as:

$$|\theta_{\text{obs}}| < \theta_{\text{range}} \quad \text{and} \quad r_i < d_{\text{critical}} \tag{14}$$

where $\theta_{\text{obs}}$ is the angle between the robot's current heading and the obstacle's position. Mathematically, the obstacle detection logic consists of calculating the distance from the robot to several objects within a given angular range and comparing those distances with $d_{\text{critical}}$. The laser scan data provides the distances in a polar coordinate system where each range $r_i$ is related to an angle $\theta_i$. If any range $r_i$ is less than $d_{\text{critical}}$, robot stops:

$$r_i < d_{\text{critical}} \quad \Delta(\text{obstacle detected}) \tag{15}$$

When the robot stops because of detection of any obstacle $\Delta$ between its path and enters a state of emergency stop. Then our proposed system try to attempt replans its path towards the target object.

### 3.7. Replanning Requests

The decision to trigger the replanning mechanism depends on two parameters, such as the number of attempts already made and a cool down period between replans. Therefore, the maximum allowed replans have not been exceeded at a limit of 5 attempts. This logic works in a condition if any obstacle is detected $\Delta$ and replan attempts are < the max replans and $(t_c - t_{lr}) > t_{cd}$.

Where, $t_c$ is the current time, $t_{lr}$ is the time of the last replanning, and $t_{cd}$ is the minimum time interval between consecutive replans. If these conditions are satisfied, the robot must change to the Replanning state, and the emergency stop flag is kept set to true until a new path has been planned. Once the robot enters the Replanning state, it requests the generation of a new set of waypoints that avoid the detected obstacle. The robot will check the environmental data and then decide which areas are clear of obstacles. In doing so, it checks where the robot is and in which direction it is heading in relation to the obstacles and recalculates the route. The new path may either be a re-routing to the neighbor way-point or altogether a new path depending on the environment constraints. Way-point selection strategy

**Algorithm 1** LLM-Based Mobile Robot Path Planning Framework

1: **Input:** User command, current position $(x_{cp}, y_{cp})$, target position $(x_{tp}, y_{tp})$
2: **Output:** Waypoints $W = \{(x_1, y_1), \ldots, (x_n, y_n)\}$
3: Generate waypoints $W \leftarrow$ LLM(User Command)
4: **for** each waypoint $W_i = (x_i, y_i)$ in $W$ **do**
5:     **if** $x_{\min} + m \leq x_i \leq x_{\max} - m$ **then**
6:         $y_{\min} + m \leq y_i \leq y_{\max} - m$
7:         Transform $W_i$ to local frame: Equation: (8)

$$\begin{bmatrix} x_{od} \\ y_{od} \end{bmatrix} = \begin{bmatrix} \cos(\theta_{ro}) & -\sin(\theta_{ro}) \\ \sin(\theta_{ro}) & \cos(\theta_{ro}) \end{bmatrix} \begin{bmatrix} x_m \\ y_m \end{bmatrix} + \begin{bmatrix} x_{ro} \\ y_{ro} \end{bmatrix}$$

8:         Move to $W_i$
9:     **else**
10:         Regenerate $W_i \leftarrow$ LLM(Regenerate)
11:     **end if**
12: **end for**
13: **while** obstacle detected **do**
14:     **if** $r_i < d_{critical}$ **then**
15:         Stop robot: $v_{linear} = 0, v_{angular} = 0$
16:         Exit path
17:     **end if**
18: **end while**
19: **if** navigation interrupted **then** Equation: (16)
20:     $\min_{\mathbf{p}} \sum_{i=1}^{n} d_i(\mathbf{p})$    subject to    $d_i(\mathbf{p}) > d_{critical}$    $\forall i$
21: **end if**
22: **while** target not reached **do**
23:     Continue path
24: **end while**

usually uses optimization algorithms in order to shorten the distance of travel and, at the same time, avoiding obstacles. Optimization can mathematically be described as:

$$\min_{\mathbf{p}} \sum_{i=1}^{n} d_i(\mathbf{p}) \quad \text{subject to} \quad d_i(\mathbf{p}) > d_{critical} \quad \forall i \tag{16}$$

where $\mathbf{p}$ is the set of way-points, $d_i(\mathbf{p})$ is the distance from the way-point $i$ to the nearest obstacle, and $n$ is the number of way-points. After each replanning trial, the system goes into a cool-down phase to prevent the robot from making too many re-plans in a short period. A paramter $t_{cd}$, here set to 5 seconds, defines this time cooldown. This is provided not to allow the robot to continuously trigger replans due to failure in successfully avoiding obstacles. If a predefined number of maximum replans occurs, the system error shall be reported or asks for a manual intervention. This retry logic is very important for preserving system stability and for excluding excessive computational load by re-planning attempts repeatedly without any success, making the system not get stuck into infinite replanning loop. During replanning, the robot's state machine follows a series of transitions. The proposed LLM-based mobile robot path planning framework is shown in Algorithm 1, which efficiently interprets user commands and generates real-time navigation paths.

## 4. Study Design

In this section we described the experimental design to comprehensively evaluate the navigational proficiency of an LLM-based autonomous agent across different environments. These experiments were conducted using a Gazebo simulator employing a Turtlebot3 Burger platform, which navigates through a series of environments. Each environment is clearly constructed to evaluate different aspects of our proposed framework, including path planning, way-point generation, and obstacles avoidance. The environments progress to difficulty levels, from a basic straight

Tariq et al.: *Preprint submitted to Elsevier*          Page 10 of 18

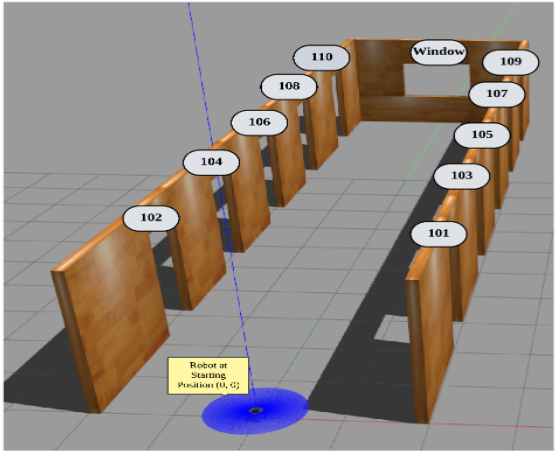 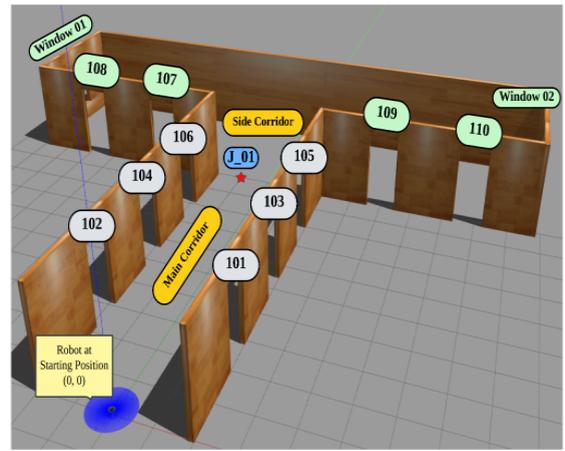

(a) Straight corridor with room number plates and a window.

(b) Main and side corridors with room numbers and two windows.

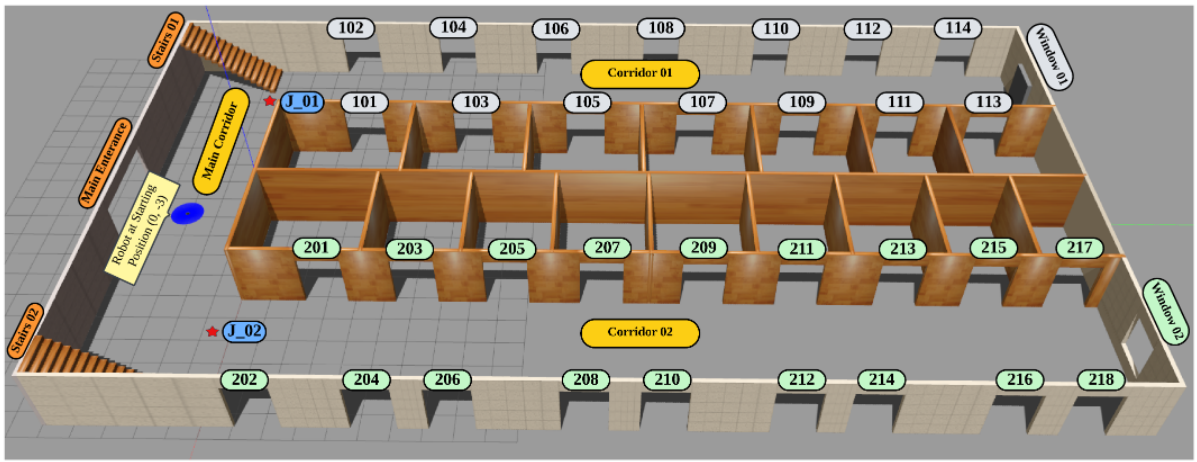

(c) U-shaped environment with three interconnected corridors, room numbers, stairs, and junction points.

**Figure 5**: Illustration of three test environments used in the study. These environments are progressively complex to evaluate the robot's navigational performance.

corridor to ones that had branching corridors and junction points. These configurations evaluate in such a way as to exercise features for dealing with narrow passages, handling junctions, and reaching targets placed in multiple locations throughout various scenarios.

### 4.1. Environment Setup and Configuration

The first environment (a) consists on single straight corridor that aims to test basic navigation capabilities of the agent in a simple linear path. It has a dimension of 15m in length and 4m in width with 11 objects serving as target locations where room number plates from 101 to 110 placed on both sides and at the end of the corridor, there is a window. The robot should be able to move autonomously from any starting position to any object in the corridor. The simplicity of this environment allows for testing basic movement capabilities in a linear space.

The next one is complex environment (b) consisting of two interconnected corridors named as main and side corridor. The main corridor is 10m in length and 4m in width, while the side corridor is 18m in length and 4m in width. In this environment, room numbers from 101 to 106 are in the main corridor and on the other hand, room numbers 107 to 110 are in side corridor. Additionally, there are two windows one on the left and the other is on the right side placed at the end of the side corridor. The difficulty in this setting then, lies in navigating through the junction point connecting the main and side corridors to get into a position from which one can navigate through two clearly separated corridors with a different arrangement of rooms. Our LLM-based path planner has to be able to generate efficient waypoints and navigate robustly to different target locations in both corridors.

The most challenging third and last environment (c) which has three interconnected corridors in a U-shaped layout. The main corridor serves as a central path with a length of 5m and a width of 18m long. Corridor 1 and Corridor 2 are on the sides each one has a length of 29.5m and a width of 5m. Corridor 1 have room numbers from 101 to 114, while Corridor 2 features numbers 201 through to 218 on the room plates. The environment also includes two staircases, located on both sides of main corridor, with one stair positioned near Corridor 1 and the other near to Corridor 2. Additionally, the layout includes two junction points that connects both corridors with the main corridor. With this large U-shaped configuration and multiple junction points, this is the most challenging environment that will let the robot handle the multitask navigation capabilities.

The design of these environments was carefully selected to evaluate the robot's performance in a range of navigation scenarios, from simple, linear navigation in the first environment to more complex junction-based navigation in the second and third environments illustrated in Fig. 5. The environments are designed to challenge the ability of our proposed approach in autonomously generating waypoints, navigating through corridors, and handling obstacles with a high degree of accuracy in reaching the target locations. For all three environments, the Turtlebot3 Burger robot is equipped with the following configuration parameters:

- **Linear Speed:** 0.4 m/s
- **Angular Speed:** 2.0 rad/s
- **Distance Threshold:** 0.1 m (waypoint reach tolerance)
- **Angle Threshold:** 0.087 rad (angular alignment tolerance)

Our proposed framework generates waypoints based on current location and target objects in the environments. The way these environments are set up allows a comprehensive analysis of the navigation capabilities of the robot under a range of scenarios of increasing complexity.

### 4.2. Testing Scenarios and Methodology

The methodology for the evaluation of our proposed framework for navigational capabilities across these three environments follows a series of structured experiments. Subsequent evaluations across all environments were tested in five separate runs, with a fixed set of commands given to the robot in order to evaluate its performance under different environmental configurations. These commands involved sequential movements from one target object to another. In the first two environments, the starting position of the robot was at the origin location of (0,0), while in environment (c), the starting position was set at (0, -3). The task for the robot was to follow the commands given to it, each directing it to a specific room number or any other object in the environment. We choose 3 different large language models (LLMs) from Ollama library including Llama3.1 (8B) Dubey et al. (2024), Qwen2.5 (7B) Yang et al. (2024), and Mathstral (7B) which is upgraded version of mistral Jiang et al. (2023), performs well on public benchmarks for generate accurate outputs based on the prompts provided by users.

These tests were conducted under controlled conditions to ensure the robot's ability to navigate autonomously. This testing methodology provided a clear and repeatable structure by which to test the path planning capabilities of the robot in an increasingly complex environment, ranging from a simple straight corridor to U-shaped ones with multiple junctions and obstacles.

### 5. Performance Metrics and Evaluation

While the previous section described the experimental setup, this part details the system performance evaluation. The experimental design provided a necessary framework for testing, but it is in the performance metrics that one can find quantifiable means to assess the success and limitations of the proposed LLM-based autonomous navigation system. Our set of performance metrics was defined as a way to measure both efficiency and robustness against adaptability in real-world scenarios as accurately as possible. Accordingly, in the following explanation, we will go in detail about the design of these performance metrics and their corresponding calculation methods, which we chose specifically to evaluate the robotic system's navigational proficiency on three environments explained in the previous section.

Tariq et al.: *Preprint submitted to Elsevier* Page 12 of 18

Our first metric is path planning time, calculates the time difference between issuing a navigation command and receiving the corresponding waypoints from the path planner. This metric helps in understanding the efficiency of the LLM model responsiveness time. Mathematically, the path planning time can be defined as the time difference between the receiving of waypoints $t_\varepsilon$ from LLM and the issuance of the command $t_\varkappa$ by the user.

$$\text{Path Planning Time} = t_\varepsilon - t_\varkappa \tag{17}$$

The next metric is execution time that measures the total time taken by the robot to execute a set of waypoints generated by LLM, from the start of movement to the completion of navigation. This metric provides critical insight into the speed of the robot and the efficacy in following the planned path. The execution time is calculated based on the difference in the time when the robot completed the navigation $t_\varphi$ and the beginning time of executing the waypoints $t_\varrho$.

$$\text{Execution Time} = t_\varphi - t_\varrho \tag{18}$$

The third metric we designed is the success rate of waypoint generation, which quantifies the reliability of LLM to process for generating waypoints. It is defined as the ratio of successful attempts at generating waypoints $\Phi$ to the total number of attempts $\Psi$. This metric will be useful in assessing the strength of LLMs in generating valid navigation waypoints with no errors. The success rate can be computed as:

$$\text{WGSR} = \left(\frac{\Phi}{\Psi}\right) \times 100 \tag{19}$$

The path length metric measures the total length of distance travelled by the robot in executing its waypoints. This is a very important metric to assess the efficiency of the robot's path - reflecting the shortest or most optimal route taken. The path length is calculated by summing the distances between consecutive waypoints on the robot's trajectory:

$$\text{Path Length} = \sum_{i=1}^{n} d_i \tag{20}$$

where, $d_i$ is the Euclidean distance between the $i$-th and $(i + 1)$-th waypoints, and $n$ is the total number of executed waypoints. Next metrics is collision detection events which count when the robot detects an obstacle within a predefined critical proximity that can activate the safety mechanism of either obstacle avoidance or replanning. This metric is of high importance because it considers how well the robot may adapt to unexpected obstacles in its path in real-time to ensure safe and reliable navigation. Mathematically, the number of collision events calculated as:

$$\text{Collision Detection Events} = \sum_{i=1}^{m} \mathbf{1}_{\{r_i < d_{\text{critical}}\}} \tag{21}$$

where, $r_i$ is the distance measurement to the obstacle at the $i$-th instance, $d_{\text{critical}}$ is the critical threshold distance for obstacle detection, and $\mathbf{1}$ is the indicator function, which is 1 if the condition is met (i.e. the robot detects an obstacle within the critical distance) and 0 otherwise. Finally, the replanning rate is the last metric that represents how often the robot must replan the path during execution due to obstacles or other unexpected circumstances. It reflects the adaptability of the robot and how well it can cope with changes in the environment. Replanning rate is determined by the ratio of replanning attempts $\mu$ to the total execution time $\Gamma$:

$$\text{Replanning Rate} = \frac{\mu}{\Gamma} \tag{22}$$

This formula gives the number of replanning attempts per unit time and can provide an indication of how often the robot needs to readjust its path to either avoid obstacles or handle unexpected changes. These six performance metrics were chosen in order to give a holistic view of the robot's navigation capabilities and enable us to examine the efficiency, reliability, and safety of the robot across various types of environments. The following will present an analysis of the results obtained from these metrics and discuss the performance of the system in detail.

Table 1
Performance Metrics for LLM Models Evaluation: This table summarises the path planning time, waypoint generation success rate, and replanning count for different navigation commands, where "SP" represents the starting point and "RNP" is the Room Number Plate.

| | | Environment (a) | | | | | | | |
|---|---|---|---|---|---|---|---|---|---|
| | llama3.1 | Mathstral | Qwen2.5 | llama3.1 | Mathstral | Qwen2.5 | llama3.1 | Mathstral | Qwen2.5 |
| Commands | Path Planning Time ↓ | | | WP-Generation Success Rate ↑ | | | Replanning Count ↓ | | |
| SP → Window | **11.67** | 18.07 | 12.55 | **100%** | **100%** | **100%** | **0** | **0** | **0** |
| Window → RNP 105 | **6.22** | 12.4 | 7.64 | - | - | - | - | - | - |
| RNP 105 → RNP 102 | 6.46 | **5.47** | 6.01 | - | - | - | - | - | - |
| RNP 102 → RNP 108 | 7.46 | **4.64** | 6.85 | - | - | - | - | - | - |
| RNP 108 → RNP 103 | 7.54 | 12.24 | **6.7** | - | - | - | - | - | - |
| | | Environment (b) | | | | | | | |
| SP → RNP 106 | 8.83 | **4.73** | 9.47 | **100%** | **100%** | **100%** | **0** | **0** | **0** |
| RNP 106 → Window 02 | 6.13 | **5.41** | 55.87 | - | - | 50% | - | - | 1 |
| Window 02 → RNP 108 | 14.74 | 6.67 | **5.9** | 33% | - | 100% | 3 | - | 0 |
| RNP 108 → RNP 101 | **6.66** | 17.39 | 7.06 | **100%** | - | - | **0** | - | - |
| RNP 101 → RNP105 | **5.82** | 5.94 | 11.1 | - | - | - | - | - | - |
| | | Environment (c) | | | | | | | |
| SP → RNP 107 | **12.71** | 13.31 | 43.2 | **100%** | **100%** | **100%** | **0** | **0** | **0** |
| RNP 107 → Main Entrance | **11.93** | * | * | - | * | * | - | * | * |
| Main Entrance → RNP 206 | **13.48** | 18.93 | 22.73 | - | 50% | 100% | - | 2 | 0 |
| RNP 206 → RNP 104 | **11.56** | * | * | - | * | * | - | * | * |
| RNP 104 → Stairs 02 | 10.36 | **5.18** | 45.28 | - | **100%** | 50% | - | **0** | 2 |

(a) *Note: Asterisk (\*) indicates a failure, and (-) indicates the same value as above.*

## 5.1. Discussion on System Performance

Our proposed LLM-based path planning system was evaluated in three environments with different complexities that are explained in the experimental setup. The results are presented in Table 1 and 2 are based on two distinct sets of metrics including LLM-specific metrics and robot navigation metrics that are outlining the robustness our proposed approach using with three LLM models form Ollama open-source library including llama3.1, mathstral, and qwen2.5. The metrics are carefully chosen in order to measure responsiveness, reliability, and adaptability of these LLM models in generating waypoints that shall enable successful navigation for a mobile robot in complex environments.

The LLM-specific metrics including path planning time, waypoint generation success rate, and replanning count offer insights into how efficiently and reliably each model was able to translate users' high-level commands and execute them into actionable navigation for mobile robots. llama3.1 show cased the best performance without any failure in path planning time across all environments. This performance underlines its efficiency of fast responses to user commands and generating waypoints in a timely manner. Mathstral shows respectable efficiency but in some cases, it slightly lagging and fails in waypoint generation, which is behind llama3.1 in more complex scenarios. In contrast, qwen2.5 showed relatively slower responses, especially in environment (c), where extended corridors and multiple junctions posed significant challenges. The LLM-specific results are summarized in the Table 1.

The execution time, path length, and collision detection events are robot navigation specific metrics those are providing a comprehensive view of the system's operational efficiency and safety during navigation. As shown in Table 2, llama3.1 demonstrated consistent execution times across all environments. This balance between efficiency and reliability underlines its suitability for navigation in complex layouts without compromising the task of completion. Mathstral executed somewhat slower with two times failure in a complex scenario. The qwen2.5 also not performed well in execution time, which pointed to its inability to navigate in complex conditions.

Another important metric in assessing navigation efficiency is path length, where again llama3.1 performed better. It generated paths that were efficient while successfully completing all tasks. Mathstral had performed well in less complex environments but in environment (c), it has two failures that showed to be less optimal in path planning, leading to longer trajectories in certain cases. Qwen2.5 has achieved the shortest path length in more cases compared to others but failed to complete some tasks in environment (c), undermining its overall performance. These results suggests that Qwen2.5 might be good in generating shorter paths, but its reliability is questionable in complex scenarios.

Table 2
Performance Metrics for Mobile Robot Navigation Evaluation: This table presents the results for evaluation of efficiency LLMs in terms of execution time and path planning with respect to collision avoidance using three models: llama3.1, Mathstral, and Qwen2.5.

| Commands | llama3.1 | Mathstral | Qwen2.5 | llama3.1 | Mathstral | Qwen2.5 | llama3.1 | Mathstral | Qwen2.5 |
| | Execution Time ↓ | | | Path Length ↓ | | | Collision Detection Event | | |
| --- | --- | --- | --- | --- | --- | --- | --- | --- | --- |
| Environment (a) | | | | | | | | | |
| SP → Window | **36.08** | 36.13 | 36.17 | **14.32** | 14.32 | 14.36 | No | No | No |
| Window → RNP 105 | 42.21 | 27.17 | **23.38** | **6.73** | 7.11 | 7.06 | - | - | - |
| RNP 105 → RNP 102 | **26.98** | 34.27 | 39.18 | 7.9 | 10.97 | **7.29** | - | - | - |
| RNP 102 → RNP 108 | **26.22** | 30.89 | 37.5 | **7.96** | 7.98 | 8.02 | - | - | - |
| RNP 108 → RNP 103 | 35.01 | 43.33 | **20.88** | 6.36 | 6.41 | **5.81** | - | - | - |
| Environment (b) | | | | | | | | | |
| SP → RNP 106 | **29.93** | 39.51 | 45.72 | 8.3 | 9.89 | **7.96** | No | No | No |
| RNP 106 → Window 02 | 43.84 | **42.38** | 47.19 | 11.49 | 12.66 | **10.76** | - | - | Yes |
| Window 02 → RNP 108 | 49 | **43.03** | 48.45 | 19.64 | **14.78** | 15.02 | Yes | - | No |
| RNP 108 → RNP 101 | **46.89** | 63.11 | 57.03 | 14.89 | 15.78 | **14.88** | No | - | - |
| RNP 101 → RNP105 | **22.59** | 26.64 | 27.12 | **5.25** | 5.71 | 6.71 | - | - | - |
| Environment (c) | | | | | | | | | |
| SP → RNP 107 | 98.76 | **73.41** | 86.89 | 32.35 | 26.09 | **22.8** | No | No | No |
| RNP 107 → Main Entrance | **84.42** | * | * | **25.72** | * | * | - | * | * |
| Main Entrance → RNP 206 | 74.23 | 182.98 | **79.12** | 21 | 68.05 | **18.28** | - | Yes | No |
| RNP 206 → RNP 104 | **100.34** | * | * | **32.76** | * | * | - | * | * |
| RNP 104 → Stairs 02 | **69.76** | 75.59 | 78.43 | **22.1** | 22.46 | 23.43 | - | No | Yes |

Table 3
Average Performance of LLM Models Across All Environments.

| Metrics | llama3.1 | Mathstral | Qwen2.5 |
| --- | --- | --- | --- |
| Path Planning Time (s) ↓ | **9.43** | 11.56 | 18.1 |
| Execution Time (s) ↓ | **52.4** | 55.2 (2*) | 48.02 (2*) |
| Path Length (m) ↓ | 15.7 | 17.09 (2*) | **12.49 (2*)** |
| WP-Generation Success Rate ↑ | **95.50%** | 83.30% | 80% |
| Replanning Count ↑ | | | |
| Collision Detection Events ↑ | **93.30%** | 80% | 73.30% |

Safety during navigation is measured by the events of collision detections, which further evaluate the performances of the LLM models. Llama3.1 did not record any collision events across all environments except for one case in environment (b) shows it is ability of generating collision-free paths and is able to respond effectively against obstacles. The plots in Fig. 6 demonstrating the results of waypoint generation and validation for the best performance runs across all environments using the llama3.1 model. These results showcased the system robustness in generating efficient and collision-free paths for navigation. While mobile robot navigating through the environment (b) and (c) using Mathstral and Qwen2.5 recorded some collision events, which is reflecting comparative challenges with ensuring safe navigation. These results again make llama3.1 a suitable model in view of applications where reliability and safety are crucial.

For waypoint generation, llama3.1 again proved to be the most reliable model, with a high success rate. This metric underlines the robustness of llama3.1 to handle all kinds of navigation commands without errors, even in the most complex environments. By contrast, Mathstral and Qwen2.5 resulted in the relatively lower success rates, including prominent failures in environment (c). These results demonstrate how much better llama3.1 is at scaling to higher environmental complexity, which is an important requirement for real-world deployments in dynamic environments. Moreover, this adaptability is further solidified in the replanning count metric, where the fewest number of replanning attempts was reported by llama3.1, whereas Mathstral and Qwen2.5 have more frequent replanning due to difficulty in recalibrating the navigation in case of any impediment that came along unexpected.

The overall averages of all metrics summarized in Table 3 show that llama3.1 consistently outperformed other models. Indeed, its well-balanced performance in all environments, along with the high success rate, very low

Tariq et al.: *Preprint submitted to Elsevier* Page 15 of 18

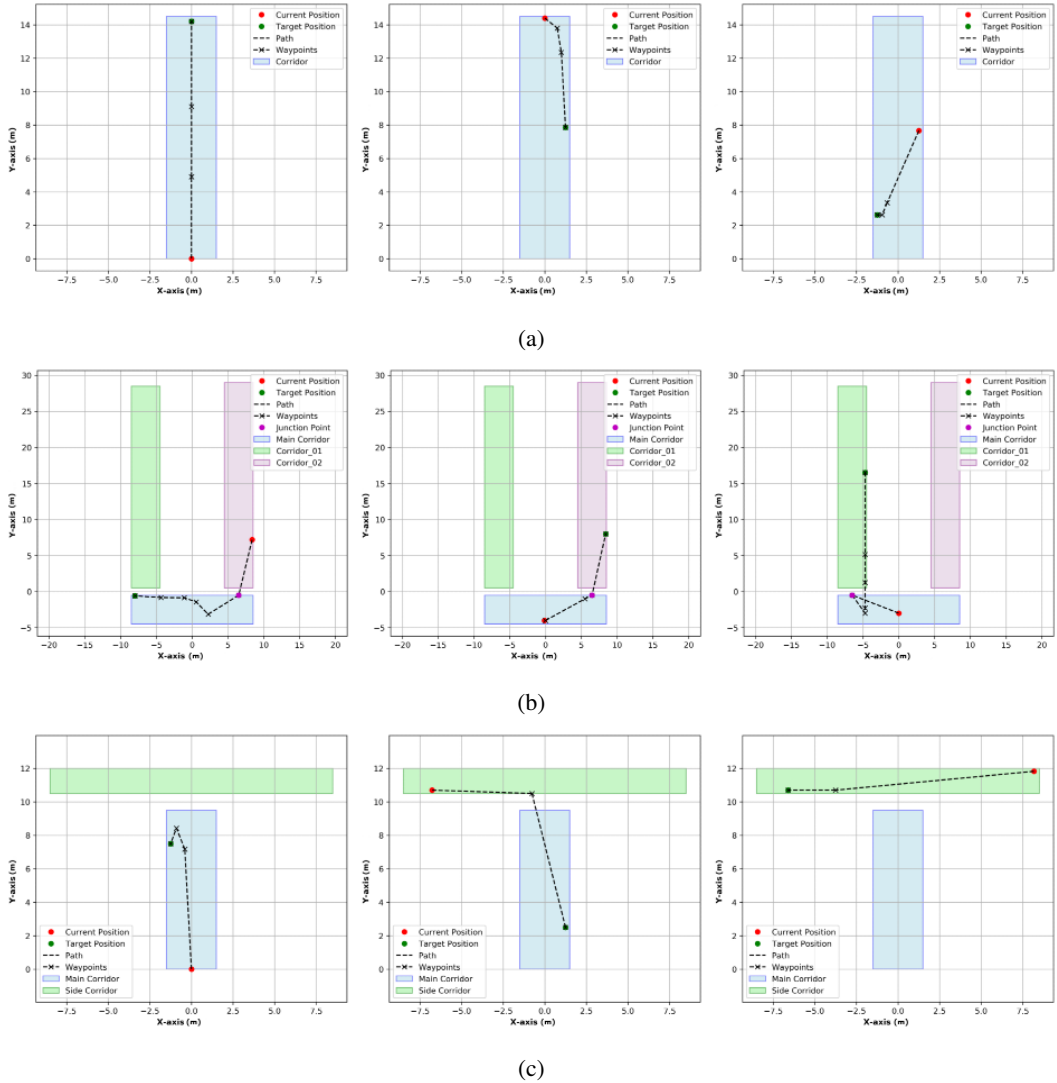

**Figure 6**: This visualization presents the waypoint generation results across three different environments. Sub-figure (a), (b), and (c) each presents the generated waypoints for one of the three environments used in this study by the llama3.1 model. These are the best results out of overall 45 runs that showcased the capability of llama3.1 model to generate optimal paths in all environments.

replanning, and absence of collision events, definitely positions it as the most robust and adaptable model for LLM-based path planning. While Mathstral and Qwen2.5 performed well in some cases but their inability to handle complex scenarios are the limitations and they are ensuring consistent reliability underscore the challenges of achieving comprehensive performance in such tasks.

These results proved the effectiveness of our proposed LLM-based path planning framework for mobile robots by using different LLM models, especially llam3.1, which showed the highest efficiency, reliability, and safety during navigation and thus can be promising for real-world applications. Its ability to balance responsiveness, accuracy, and adaptability sets a strong precedent for the integration of LLMs in autonomous navigation systems.

### 5.2. Potential Failure Modes and Mitigation Strategies

The challenges observed during the implementation and evaluation of our proposed LLM-based path planning systems and ways to mitigate them are outlined. A major challenge was that LLM models, like Mathstral and Qwen2.5, sometimes fail to give valid waypoints in complex scenarios like environment (c). Most of these failures were because of inability to correctly interpret spatial constraints or to handle ambiguous situations. Fine-tuning of LLMs prompts and incorporating a more robust waypoint generation mechanism can improve their accuracy and reliability.

There were also execution inconsistencies reported, especially in environment (b), where Qwen2.5 showed increased collision detections and replanning requirements, pointing a gap in obstacle detection and motion control. Frequent replanning, especially in complex layouts, has also been one of the challenges. High replanning rates not only extend the navigation time but also increase computational loads. Proactive replanning strategies that precompute alternative paths, and adaptive thresholds for triggering replanning events, could alleviate these inefficiencies.

Most limitations are Model-specific because all LLM models are trained for a specific task. As we observed that llama3.1 excelled in efficiency and reliability, while Mathstral showed competitive execution times in certain scenarios but struggled with complexity. Qwen2.5 delivered shorter path lengths in successful events but faced adaptability challenges. Our future work will be on improvements for LLM-based path planning approaches would be in developing a hybrid system. It might achieve higher adaptability and accuracy of generated waypoints in complex conditions, enabled by the methods such as Chain of Thought, being a well-established strategy to improve the reasoning capabilities of LLMs. The goal will be to further create an enhanced system that will easily tackle different challenges in navigation by combining several LLM models, each of which performs exemplary well on certain tasks. The hybrid approach can also be a way out to fix some context-specific limitations identified in individual models like Mathstral and Qwen2.5 for more reliable and scalable solutions of real-world navigation.

## 6. Conclusion

In this article, we have proposed an LLM-based path planning framework for mobile robots. We have showed that with the natural language processing of LLM, robots can effectively understand human commands and then translate these inputs into executable navigation tasks, furthermore it has the ability to dynamically adjust the path in case of any obstacle detected during navigation, while offering reliable performance across a wide range of scenarios. Our simulation experiments verify that the proposed framework is effective in three progressively complex environments, with quantified metrics such as path planning time, waypoint generation success rate, execution time, and collision detection. While these simulation results are promising, the next step will be focused on the performing real-world experiments that would prove the robustness of this system in dynamic settings. The future work will mainly be enhancing the framework with a hybrid LLM system that would improve adaptability and accuracy, address the limitations presented in this paper, and pave the way for practical deployment in autonomous navigation through diverse real-world applications.

## Acknowledgments

The research was supported by the National Natural Science Foundation of China(No. 62150410441),the National Natural Science Foundation of China (No. 62372232), the Fundamental Research Funds for the Central Universities (No. NG2023005),the collaborative Innovation Center of Novel Software Technology and Industrialization.